\documentclass[sigconf]{acmart}
\usepackage{balance}
\usepackage{algorithmic}

\usepackage[linesnumbered,ruled,vlined]{algorithm2e}

\AtBeginDocument{%
  }

\setcopyright{cc}

\copyrightyear{2026}
\acmYear{2026}
\setcopyright{cc}
\setcctype{by-nc-nd}
\acmConference[ACMSE 2026]{2026 ACM Southeast Conference}{April 23--25, 2026}{Troy, AL, USA}
\acmBooktitle{2026 ACM Southeast Conference (ACMSE 2026), April 23--25, 2026, Troy, AL, USA}
\acmDOI{10.1145/3746467.3801496}
\acmISBN{979-8-4007-2062-8/2026/04}

\begin{document}
\pagestyle{empty}

\title{No Pedestrian Left Behind: Real-Time Detection and Tracking of Vulnerable Road Users for Adaptive Traffic Signal Control}

\author{Anas Gamal Aly}
\affiliation{%
  \institution{Stetson University}
  \city{DeLand}
  \state{Florida}
  \country{USA}
}
\email{agamal@stetson.edu}

\author{Hala ElAarag}
\affiliation{%
  \institution{Stetson University}
  \city{DeLand}
  \state{Florida}
  \country{USA}
}
\email{helaarag@stetson.edu}

\renewcommand{\shortauthors}{Aly et al.}

\begin{abstract}
Current pedestrian crossing signals operate on fixed timing without adjustment to pedestrian behavior, which can leave vulnerable road users (VRUs) such as the elderly, disabled, or distracted pedestrians stranded when the light changes. We introduce \textbf{No Pedestrian Left Behind (NPLB)}, a real-time adaptive traffic signal system that monitors VRUs in crosswalks and automatically extends signal timing when needed. We evaluated five State-of-the-Art object detection models on the BG Vulnerable Pedestrian (BGVP) dataset, with You Only Look Once (YOLO) v12 achieving the highest mean Average Precision at 50\% (mAP@0.5) of 0.756. NPLB integrates our fine-tuned YOLOv12 with ByteTrack multi-object tracking and an adaptive controller that extends pedestrian phases when remaining time falls below a critical threshold. Through 10,000 Monte Carlo simulations, we demonstrate that NPLB improves VRU safety by 71.4\%, reducing stranding rates from 9.10\% to 2.60\%, while requiring signal extensions in only 12.1\% of crossing cycles.
\end{abstract}

\ccsdesc[500]{Computing methodologies~Object detection}
\ccsdesc[500]{Computer systems organization~Embedded systems}
\ccsdesc[500]{Computing methodologies~Tracking}
\ccsdesc[300]{Applied computing~Transportation}
\ccsdesc[300]{Mathematics of computing~Markov-chain Monte Carlo methods}

\keywords{pedestrian detection, vulnerable road users, adaptive traffic signal control, computer vision, object tracking, AI for social good, real-time systems}

\maketitle

\section{Introduction}

Pedestrian safety remains a critical challenge for modern cities. The World Health Organization estimates that more than 270,000 pedestrians die annually in road traffic crashes, representing nearly 22\% of global road fatalities \cite{WHO2023Manual}. Urban intersections are particularly risky because fixed-time pedestrian phases assume uniform walking speeds; in reality, speeds vary substantially across individuals and are often lower for elderly, disabled, or visually impaired pedestrians. This creates a risk that some pedestrians are still in the crosswalk when the signal expires. Evaluations of mid-block pedestrian signals (MPS) have shown measurable reductions in pedestrian--vehicle conflicts, reinforcing that timing aligned to observed behavior can improve safety \cite{Ahsan2025AAnP}.

Abdelrahman et al. \cite{Abdelrahman2025} introduced Video-to-Text Pedestrian Monitoring (VTPM), a Large Language Model (LLM) framework that generates real-time reports with 0.33 second latency. There are various machine learning methods that could be applied to Intelligent Transportation Systems (ITS) ~\cite{gangwani2021applications}. For traffic-flow detection, methods such as  k-nearest Neighbor (k-NN), support vector regression (SVR), SVM, or long short-term memory (LSTM) were used. For accident detection and prevention, methods such as regression tree (CART, k-NN) ~\cite{gangwani2021applications} were used. Mohandas et al. \cite{MOHANDAS2019101499} defines smart cities as the establishment of all the facilities required in the context of demand-based needs.

Machine learning methods such as Neural Networks have been found to be useful in improving the energy efficiency of street lights and creating weather-aware methods~\cite{burgos2012improving, kolasa2016concept, 8342746}. To implement smart city lights, Kama et al. ~\cite{8285797} introduced solar street lights that collect data in real time on an online server to help plan the cleaning of the solar panels.

In this work, we present NPLB (No Pedestrian Left Behind), a real-time adaptive  traffic signal system that uses computer vision to detect and track vulnerable road users at crosswalks and automatically extend pedestrian signal phases when needed.  We benchmark five State-of-the-Art object detection models on the BG Vulnerable  Pedestrian (BGVP) dataset to identify the best-performing architecture for VRU  detection, and evaluate the full NPLB system through 10,000 Monte Carlo simulations  that incorporate realistic detection failure rates. Our results demonstrate that NPLB reduces pedestrian stranding rates by 71.4\% while requiring signal extensions in  only 12.1\% of crossing cycles, establishing the viability of vision-based adaptive signal control as an equity-focused safety intervention.

\section{Related Works}

\paragraph{Adaptive Signal Control}
Traditional practice in pedestrian signal design relies on assumed walking speeds, which often under-serve slower pedestrians. Muley et al. \cite{Muley2017Procedia} reviewed global practices and highlighted the limitations of fixed-time approaches. Adaptive signal control strategies have been shown to improve multimodal performance; for example, Akyol et al. \cite{AKYOL2020704} demonstrated that adjusting green times dynamically based on demand improved both throughput and pedestrian safety. Field evaluations of mid-block pedestrian signals (MPS) report reduced probabilities of moderate and serious conflicts in before-and-after studies \cite{Ahsan2025AAnP}. Networked sensor-based adaptive controllers also show potential for scaling adaptive strategies across corridors and cities \cite{Wang2025Sensors}. Deploying Split Cycle Offset Optimisation Technique (SCOOT) algorithm for adaptive signal timings improved the pedestrian travel times by 30\% while increasing vehicle delays by 11\% ~\cite{AKYOL2020704}

\paragraph{Prediction Models for Pedestrian Intention}
Several recent studies move beyond detection to forecasting pedestrian behavior. Abdelrahman et al. \cite{Abdelrahman2025VRUCrossSafe} introduced VRUCrossSafe, a framework that predicts the crossing intention of pedestrians, cyclists, and scooter riders in real time, achieving high accuracy and inference speed. Such predictive models highlight the potential of CV-based monitoring but stop short of integrating predictions into traffic control. Our work builds on this literature by coupling real-time perception with a rule-based controller that dynamically extends pedestrian phases, addressing an important research gap.

YOLO (You Only Look Once) models offer fast, accurate pedestrian detection, while trackers such as DeepSORT~\cite{Wojke2017DeepSORT}, OC-SORT~\cite{Cao2023OCSORT} and ByteTrack~\cite{zhang2022bytetrack} maintain consistent pedestrian IDs across frames. Together, these tools allow estimation of walking speed and crossing progress directly from video feeds. Beyond detection, recent works have applied CV to predict pedestrian crossing intention, such as VRUCrossSafe, which achieved 94.7\% accuracy at real-time speed by leveraging pose and appearance features \cite{Abdelrahman2025VRUCrossSafe}.

Despite this progress, much of the literature optimizes signals primarily for vehicles or focuses on predicting pedestrian behavior without adapting the signal timing itself. Studies on adaptive signal control emphasize the benefits for traffic throughput and delay reduction, but few directly extend pedestrian crossing time based on real-time trajectory data. Moreover, vulnerable road users (VRUs), such as wheelchair users or visually impaired pedestrians, remain underrepresented in such studies \cite{AKYOL2020704, Muley2017Procedia, FHWA2021ASCT, Wang2025Sensors}. To address this gap, we introduce the following contributions:

\begin{enumerate}
    \item \textbf{NPLB: An Adaptive Pedestrian-Safety System.}
    A real-time traffic signal architecture (Figure~\ref{alg:nplb}) that integrates a fine-tuned VRU YOLOv12 detector with an adaptive signal controller to dynamically extend pedestrian crossing times based on the detection of VRUs.

    \item \textbf{Comprehensive Benchmarking Of VRU Detection Models.}
    We conduct a comparative evaluation of State-of-the-Art object detectors (YOLOv5, YOLOv11, YOLOv12, SSDLite, Fas.ter R-CNN) on the BGVP dataset, establishing performance baselines for VRU detection.

    \item \textbf{State-of-the-Art YOLOv12 Model For VRU Detection.}
    We fine-tune YOLOv12 for VRU detection and achieve an mAP@0.5 of 0.756, outperforming all tested models and demonstrating its suitability for real-time integration within NPLB due to its speed and accuracy.
\end{enumerate}

\section{NPLB (No Pedestrian Left Behind) System}
\begin{figure}[h!]
  \centering
  \includegraphics[width=0.9\linewidth]{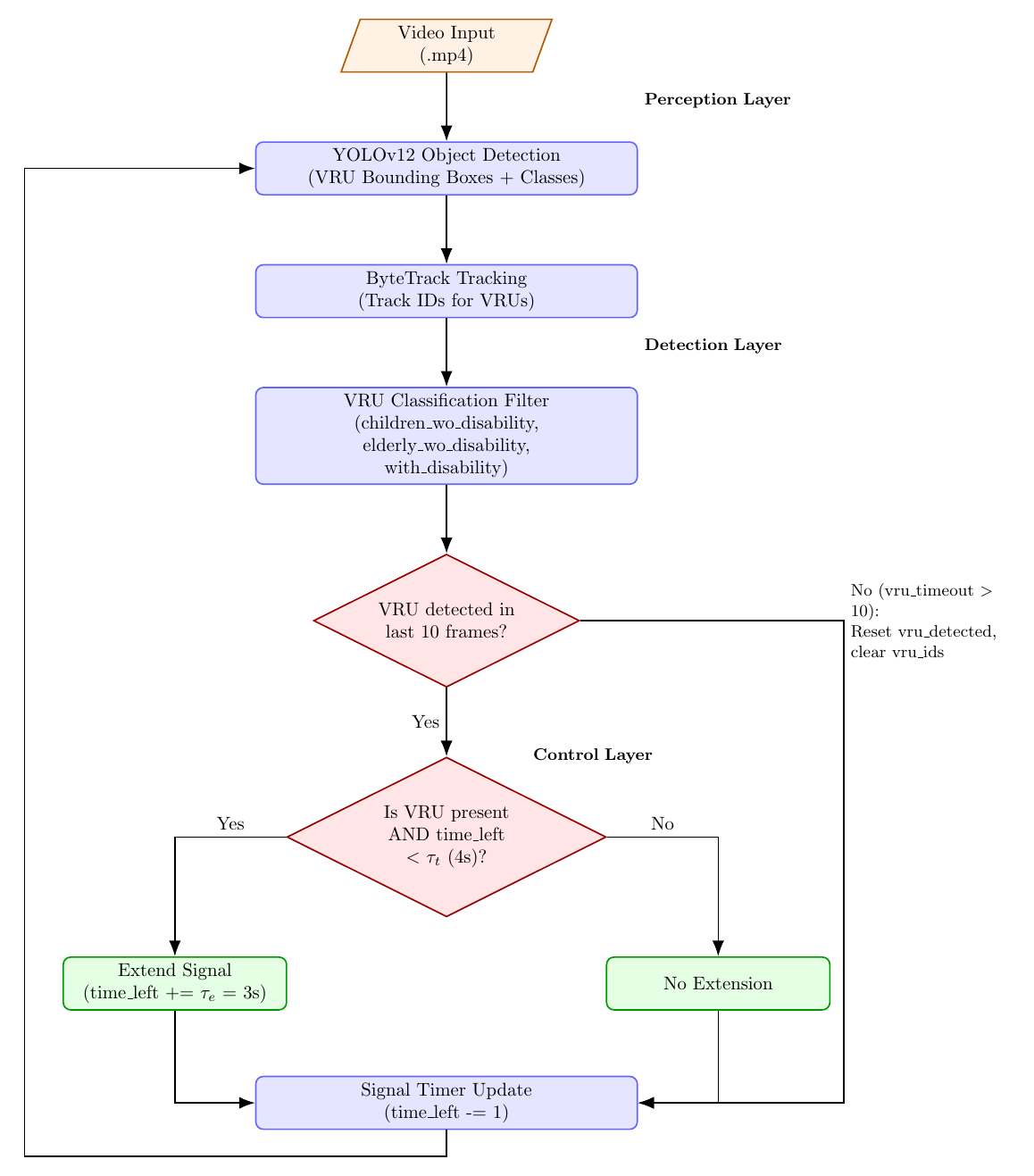}
  \caption{NPLB Architecture}
  \label{nplb-arc}
\end{figure}

Our research explores an approach to improving pedestrian safety through computer vision technology. We focus on enhancing detection capabilities specifically for vulnerable road users and implement proper mitigation through our NPLB system.

The NPLB system addresses a critical gap in pedestrian safety by dynamically adapting traffic signal timing based on real-time detection of vulnerable road users. The system architecture, illustrated in Figure ~\ref{nplb-arc}, consists of three integrated layers: perception, detection, and control.

\subsection{System Architecture}
\begin{enumerate}

\item \textbf{Perception Layer} processes video input through our fine-tuned YOLOv12 model, which detects VRUs and generates bounding boxes with class labels for each frame. These detections are then passed to \texttt{ByteTrack}, a multi-object tracking algorithm that maintains consistent track IDs across frames, enabling the system to follow individual pedestrians throughout their crossing journey.

\item \textbf{Detection Layer} filters the tracked objects to identify VRUs specifically, focusing on three vulnerability categories: children without disabilities, elderly individuals without disabilities, and persons with disabilities. This layer maintains a set of unique VRU track IDs and implements a timeout mechanism to handle temporary occlusions or detection gaps. If no VRU is detected for 10 consecutive frames (approximately 0.33 seconds at 30 fps), the system resets its VRU tracking state to prevent false positives from stale detections.

\item \textbf{Control Layer} implements the adaptive signal timing logic. When a VRU is detected in the crosswalk and the remaining signal time falls below a $\tau_t$ critical threshold (4 seconds), the system automatically extends the pedestrian phase by a $\tau_e$ predefined extension time (3 seconds). This threshold is based on typical walking speeds and crosswalk widths, ensuring that pedestrians who have already committed to crossing have sufficient time to complete their journey safely.
\end{enumerate}
\subsection{Assumptions and Scope}

The current NPLB implementation assumes
\begin{enumerate}
    \item Pedestrians are crossing legally at designated crosswalks during pedestrian signal phases
    \item Pedestrians are crossing during pedestrian signal phases
    \item An environment with pre-existing clearly marked crosswalks
    \item Adequate lighting conditions for computer vision processing.
\end{enumerate}

Illegal crossings or jaywalking scenarios require different detection and response strategies and are outside the scope of this work.

The extension time parameter can be calibrated based on local intersection characteristics, including crosswalk width, typical VRU walking speeds, and traffic patterns. We use a $\tau_e$ extension of 3 seconds.

\begin{algorithm}
\caption{No Pedestrian Left Behind (NPLB)}
\label{alg:nplb}
\begin{algorithmic}[1]
\STATE \textbf{Input:} video\_stream, initial\_signal\_time
\STATE \textbf{Parameters:} $\tau_e = 3.0$s (extension\_time), $\tau_t = 4.0$s (threshold\_time), $n_{max} = 2$ (max\_extensions)

\STATE model $\leftarrow$ YOLOv12.load("weights.pt")
\STATE vru\_ids $\leftarrow$ $\emptyset$ \COMMENT{Set of unique VRU track IDs}
\STATE vru\_detected $\leftarrow$ False
\STATE vru\_timeout $\leftarrow$ 0
\STATE time\_left $\leftarrow$ initial\_signal\_time
\STATE extension\_count $\leftarrow$ 0

\FOR{each frame in video\_stream.track(tracker="ByteTrack")}
    \STATE detections $\leftarrow$ frame.boxes
    \STATE vru\_detected $\leftarrow$ False \COMMENT{Reset per frame}
    
    \FOR{each box in detections}
        \STATE cls $\leftarrow$ box.class
        \STATE track\_id $\leftarrow$ box.id
        
        \IF{cls $\in$ \{children\_wo\_disability, elderly\_wo\_disability, with\_disability\}}
            \STATE vru\_detected $\leftarrow$ True
            \IF{track\_id $\neq -1$ and track\_id $\notin$ vru\_ids}
                \STATE vru\_ids $\leftarrow$ vru\_ids $\cup$ \{track\_id\}
            \ENDIF
        \ENDIF
    \ENDFOR
    
    \IF{not vru\_detected}
        \STATE vru\_timeout $\leftarrow$ vru\_timeout + 1
        \IF{vru\_timeout $>$ 10}
            \STATE vru\_detected $\leftarrow$ False
            \STATE vru\_ids $\leftarrow$ $\emptyset$
            \STATE vru\_timeout $\leftarrow$ 0
        \ENDIF
    \ELSE
        \STATE vru\_timeout $\leftarrow$ 0 \COMMENT{Reset timeout}
    \ENDIF
    
    \IF{vru\_detected and time\_left $< \tau_t$ and extension\_count $< n_{max}$}
        \STATE time\_left $\leftarrow$ time\_left + $\tau_e$
        \STATE extension\_count $\leftarrow$ extension\_count + 1
        \STATE \textbf{output} "EXTEND\_SIGNAL"
    \ENDIF
    
    \STATE time\_left $\leftarrow$ time\_left - 1
\ENDFOR

\end{algorithmic}
\end{algorithm}

\subsection{Operational Logic}

Algorithm~\ref{alg:nplb} describes the complete NPLB workflow. The system operates on a per-frame basis, processing each video frame through the YOLOv12 detector with ByteTrack integration. For each detected bounding box, the system checks whether the object belongs to one of the three VRU classes. When a VRU is identified, its track ID is added to the active VRU set, and the timeout counter is reset.

The decision to extend signal timing is governed by two conditions:
\begin{enumerate}
    \item At least one VRU must be actively detected in the current frame or have been detected within the last 10 frames
    \item The remaining signal time must be less than the critical threshold.
\end{enumerate}
When both conditions are satisfied, the system issues a command indicated as \texttt{EXTEND\_SIGNAL} and increments the remaining time by the extension duration.

This approach ensures that the system only extends signals when necessary, balancing pedestrian safety with traffic flow efficiency. The timeout mechanism prevents indefinite signal extensions due to detection artifacts or pedestrians who have already cleared the crosswalk, while the track ID maintenance allows the system to recognize returning or persistently present VRUs.

\section{Experiment Design and Evaluation}

\subsection{NPLB Evaluation}

To evaluate the safety impact of the NPLB system, we conducted Monte Carlo simulations comparing fixed-time signals against NPLB control. Our simulation framework models pedestrian crossing scenarios with varying crosswalk widths, and walking speeds based on pedestrian types.

Our evaluation is carried out in two phases. First, we benchmark five object detection models on the BGVP dataset to select the best-performing computer vision model. Second, we evaluate the NPLB system through Monte Carlo simulation, which  incorporates the best computer vision model's empirical recall as a parameter ($\alpha = 1 - \text{recall} = 0.26$) to realistically model per-frame missed detections without requiring a labeled video dataset.

\subsubsection{Simulation Framework}

We simulated 10,000 crossing scenarios using pedestrian walking speeds from U.S. Department of Transportation guidelines~\cite{a2016_lesson}: average adults (4.00 ft/s), wheelchair users (3.55 ft/s), and elderly adults (2.80 ft/s). Pedestrian type distribution was based on 2023-2024 U.S. Census data~\cite{a2024_secretary, bureau_2025_older}: 18\% elderly, 1.6\% wheelchair users, and 80.4\% general adults. Consistent with recent city-wide measurements reported by Moran and Laefer~\cite{Moran19092024}, crosswalk lengths in our study fall within a range of 30 to 60 feet. Longer crossings exceeding 50–60 feet are associated with higher pedestrian-collision risk~\cite{Moran19092024}.

Each simulation trial proceeds as follows. A pedestrian is assigned a type (general adult, elderly, or wheelchair user). Then according to U.S. Census proportions, a walking speed is drawn from the corresponding distribution. A crosswalk length sampled uniformly from 30-60 feet, and a randomized entry delay of 0--3 seconds to model natural crossing behavior. The fixed-time signal duration is computed using the Colorado Department of Transportation (CDOT) formula~\cite{a2024_traffic} formula for that pedestrian's assigned speed and crosswalk length.

For the NPLB condition, the adaptive controller monitors the pedestrian's estimated position frame-by-frame: at each simulated frame, the detector either successfully identifies the VRU (with probability $1 - \alpha = \text{recall}$) or misses the detection (with probability $\alpha = 0.26$). When a VRU is detected and remaining signal time falls below $\tau_t$, the controller issues an extension. A pedestrian is considered \textit{stranded} if they remain in the crosswalk when the signal phase expires. Stranding rates are computed as the fraction of stranded pedestrians across all 10,000 trials. Monte Carlo simulation has proven effective in pedestrian safety research~\cite{bagabaldo2025improving, Huang18082017}, allowing comprehensive exploration of parameter space and edge cases.
\subsubsection{NPLB System Parameters}

The NPLB system operates with the parameters in Table~\ref{tab:symbols}:

\begin{table}[h!]
\centering
\caption{System Parameters}
\label{tab:symbols}
\begin{tabular}{clcl}
\hline
\textbf{Symbol} & \textbf{Parameter} & \textbf{Value} & \textbf{Description} \\
\hline
$\rho$ & Detection Rate & 0.756 & mAP@0.5 \\
$\tau_e$ & Extension Time & 3.0 s & Extension duration \\
$\tau_t$ & Threshold Time & 4.0 s & Time-remaining trigger \\
$n_{max}$ & Max Extensions & 2 & Allowed per crossing \\
$\alpha$ & Systemic Failure & 0.26 & 1 - recall \\
\hline
\end{tabular}
\end{table}
$\tau_e$ and $\tau_t$ parameters were selected through sensitivity analysis to balance pedestrian safety with minimal traffic flow disruption.

\subsubsection{Baseline Comparison}

For the fixed-time baseline, we calculated signal durations using the Colorado Department of Transportation formula~\cite{a2024_traffic}: signal time = (crossing distance / walking speed) + buffer, where buffer $\geq$ 2 seconds. We used a 5-second buffer, representing conservative engineering practices.

\subsubsection{Parameter Optimization}

\begin{figure}[h!]
  \centering
  \includegraphics[width=0.7\linewidth]{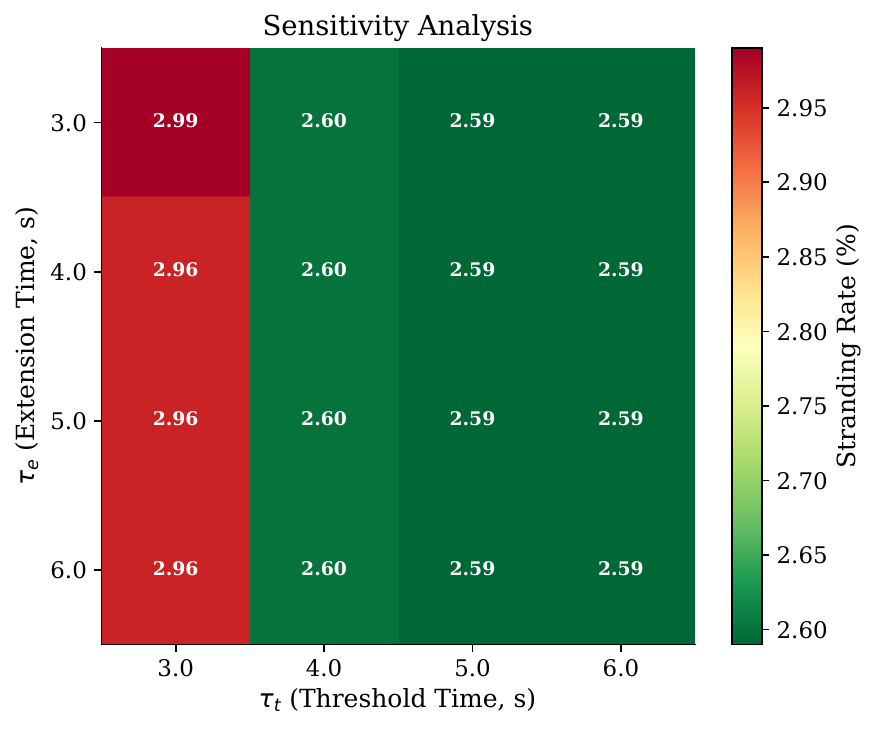}
  \caption{Heatmap of Parameter Sweep Showing Stranding Rates Across Different Combinations of Extension Time ($\tau_e$) and Threshold Time ($\tau_t$)}
  \label{fig:parameter_sweep}
\end{figure}

We conducted a parameter sweep across extension times (3.0-6.0 s) and threshold times (3.0-6.0 s) to identify optimal values (Figure~\ref{fig:parameter_sweep}). The analysis revealed that $\tau_e = 3.0$ seconds provides sufficient protection while minimizing signal extensions, and $\tau_t = 4.0$ seconds offers the most significant stranding reduction without excessive impact on total signal duration. These values represent the minimum effective intervention that maintains safety benefits.

\subsection{Model Evaluation}
Mean average precision (mAP)~\cite{schlosser2024consolidated} provides a meaningful evaluation over multiple classes for Computer Vision models. We describe it mathematically as shown in Equation ~\ref{eq:map}:

\begin{equation}
mAP = \frac{1}{n} \cdot \sum_{i=1}^{n} AP_i
\label{eq:map}
\end{equation}

\textbf{Intersection over Union (IoU)} measures the overlap between a predicted bounding box and the ground-truth box, defined as the area of intersection divided by the area of their union. Higher IoU values indicate more accurate localization of the detected object.

mAP@0.5 considers a detection correct if the predicted bounding box overlaps the ground truth by at least 50\% IoU, providing a lenient measure of detection. On the other hand, mAP@[0.5:0.95] averages the mean Average Precision over IoU thresholds from 0.5 to 0.95 in increments of 0.05, providing a stricter and more comprehensive evaluation that accounts for both detection and localization accuracy across multiple overlap levels.

\subsection{Data Collection}

After examining different pedestrian datasets including Pedestrian Intention Estimation (PIE) \cite{Rasouli2019PIE}, Joint Attention in Autonomous Driving (JAAD) \cite{Kotseruba2016JAAD}, and Caltech Pedestrians \cite{dollar_wojek_schiele_perona_2009}, we found that these datasets do not contain VRUs. We selected the BGVP dataset \cite{sharma2022comparison} because it is, to our knowledge, the only publicly available annotated dataset containing the VRU demographics targeted by NPLB: elderly pedestrians, wheelchair users, and children. Widely used video datasets such as PIE~\cite{Rasouli2019PIE} and JAAD~\cite{Kotseruba2016JAAD} do not include these labels.

The BGVP dataset follows the COCO~\cite{lin2015microsoftcococommonobjects} dataset format. To use the BGVP dataset with our YOLO models, we employ Algorithm~\ref{alg:bgvp_preprocessing} to convert the BGVP dataset format into YOLO format.

\begin{algorithm}
\caption{BGVP Dataset Preprocessing}
\label{alg:bgvp_preprocessing}
\begin{algorithmic}[1]
\STATE \textbf{Input:} BGVP dataset with split archives and COCO annotations
\STATE \textbf{Output:} YOLO-formatted dataset

\STATE Create dataset configuration with 4 VRU classes

\FOR{each split in \{train, validation, test\}}
    \STATE Extract images from archive to organized directory structure
\ENDFOR

\FOR{each split in \{train, validation, test\}}
    \STATE Load COCO annotations for split
    \FOR{each annotated image}
        \STATE Convert bounding boxes: YOLO format $\rightarrow$ COCO format
        \STATE Normalize coordinates by image dimensions
        \STATE Write YOLO label file with class and bbox per object
    \ENDFOR
\ENDFOR

\STATE Validate image and label counts match across all splits

\RETURN Preprocessed dataset ready for YOLO training
\end{algorithmic}
\end{algorithm}

\section{Model Implementation}

To identify the most effective architecture for VRU detection, we evaluate 5 State-of-the-Art object detection models fine-tuned on the BGVP dataset. Our evaluation includes three variants from the YOLO family (YOLOv5, YOLOv11, YOLOv12), known for real-time performance, alongside Faster R-CNN and SSDLite, which represent two-stage and lightweight detection approaches, respectively.

\subsection{YOLOv5, YOLOv11, YOLOv12}

For real-time VRU detection, we fine-tuned three YOLO object-detection models (v5, v11, v12) on the BGVP dataset. The YOLO models are single-stage detectors designed for speed and accuracy, and output bounding boxes, objectness scores, and class probabilities in a unified framework.

\paragraph{Training Procedure}  
We initialized each model with pre-trained weights provided by \texttt{Ultralytics} and trained on the BGVP dataset with 3 VRU classes. Training parameters were set to ensure sufficient learning while mitigating overfitting:  

\begin{itemize}
    \item \textbf{Image Size:} 640 $\times$ 640 pixels  
    \item \textbf{Batch Size:} 16  
    \item \textbf{Epochs:} 500, with early stopping patience of 50 epochs  
    \item \textbf{Data Augmentation:} horizontal flips, mosaic augmentation, small geometric transformations (rotation $\pm 10^\circ$, translation up to 10\%, scaling up to 50\%), and mild color augmentations (hue $\pm$ 0.015, saturation $\pm$ 0.7, brightness $\pm$ 0.4)
\end{itemize}

\paragraph{Implementation}  
Model training was performed using a Python library called \texttt{ultralytics}. We monitor validation performance via the mean Average Precision (mAP) metric, saving the best weights automatically. Augmentation strategies, such as mosaic and horizontal flips, were included to improve generalization given the limited size of the VRU-specific dataset.  

\subsection{Faster R-CNN}

For comparison with real-time YOLO models, we fine-tuned a Faster R-CNN model with a ResNet-50 backbone on the BGVP dataset. Faster R-CNN is a two-stage detector that first generates region proposals and then classifies each proposal while refining bounding box coordinates, offering high accuracy at the expense of inference speed.

\paragraph{Training Procedure}  
The model is initialized with weights pre-trained on the COCO dataset and trained to detect the same 3 VRU classes. Key training parameters include:

\begin{itemize}
    \item \textbf{Backbone:} ResNet-50 with Feature Pyramid Network (FPN)  
    \item \textbf{Batch Size:} 2
    \item \textbf{Epochs:} 100, with early stopping patience of 10 epochs  
    \item \textbf{Optimizer:} Stochastic Gradient Descent (SGD) with learning rate 0.005, momentum 0.9, weight decay 0.0005  
    \item \textbf{Learning Rate Scheduler:} StepLR with step size 3 and decay factor 0.1  
    \item \textbf{Data Augmentation:} Random horizontal flips (50\%) and normalization
\end{itemize}

\paragraph{Implementation}  
Training is performed using PyTorch and the TorchVision library. We monitor validation performance using mean Average Precision (mAP) and save the best-performing model weights automatically. Faster R-CNN is computationally more expensive than YOLO, so fewer epochs are used to balance training time and accuracy.

\subsection{SSDLite}
To explore lightweight detection architectures suitable for mobile and edge deployment, we fine-tuned an SSDLite320 model with a MobileNetV3-Large backbone on the BGVP dataset. SSDLite is a single-stage detector that uses depthwise separable convolutions to significantly reduce computational cost while maintaining competitive accuracy, making it well-suited for resource-constrained environments.

\paragraph{Training Procedure}  
The model is initialized with weights pre-trained on the COCO dataset and fine-tuned to detect 3 VRU classes plus background (5 classes total). We replaced the original classification head with a new \texttt{SSDLiteClassificationHead} configured for our target classes while preserving the backbone and feature extraction layers. Key training parameters include:
\begin{itemize}
    \item \textbf{Input Resolution:} 320 $\times$ 320 pixels  
    \item \textbf{Backbone:} MobileNetV3-Large with depthwise separable convolutions  
    \item \textbf{Batch Size:} 2  
    \item \textbf{Epochs:} 150, with early stopping patience of 30 epochs  
    \item \textbf{Optimizer:} AdamW with learning rate 0.001 and weight decay 0.0005  
    \item \textbf{Learning Rate Scheduler:} StepLR with step size 3 and decay factor 0.1  
\end{itemize}

\paragraph{Implementation}  
Training is performed using Python libraries called \texttt{PyTorch} and \texttt{TorchVision}. We monitor validation performance using mean Average Precision (mAP). Category IDs from the COCO-format annotations are remapped to 1-indexed labels to ensure compatibility with the model's classification head, where background is implicitly class 0.

\section{Results}

\subsection{NPLB}

\begin{figure}[h!]
  \centering
  \includegraphics[width=0.7\linewidth]{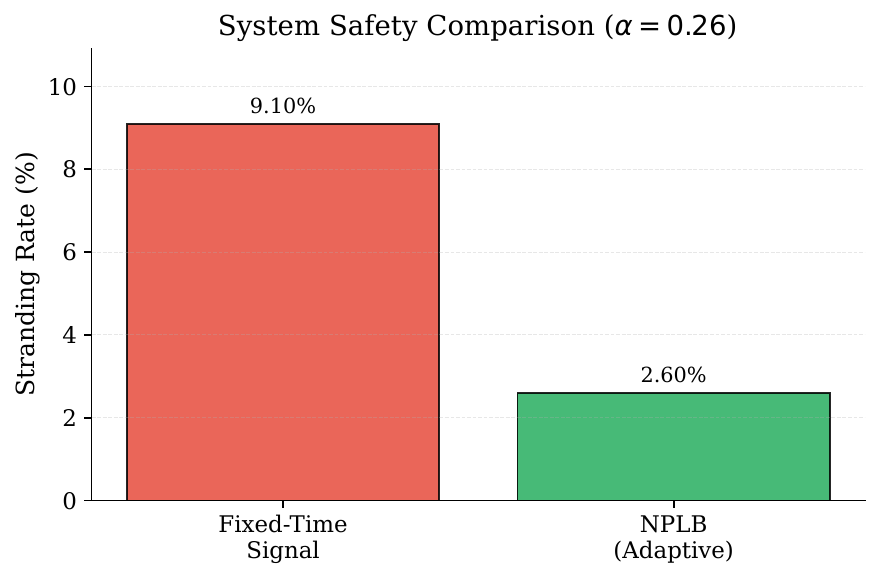}
  \caption{Comparison of Stranding Rates Between Fixed-Time Signal and NPLB Adaptive Control}
  \label{fig:stranding_comparison}
\end{figure}

Figure~\ref{fig:stranding_comparison} demonstrates that NPLB achieved a 71.4\% improvement in stranding rates compared to fixed-time signals — reducing stranding rates from 9.10\% to 2.60\%. 

\begin{figure}[h!]
  \centering
  \includegraphics[width=0.7\linewidth]{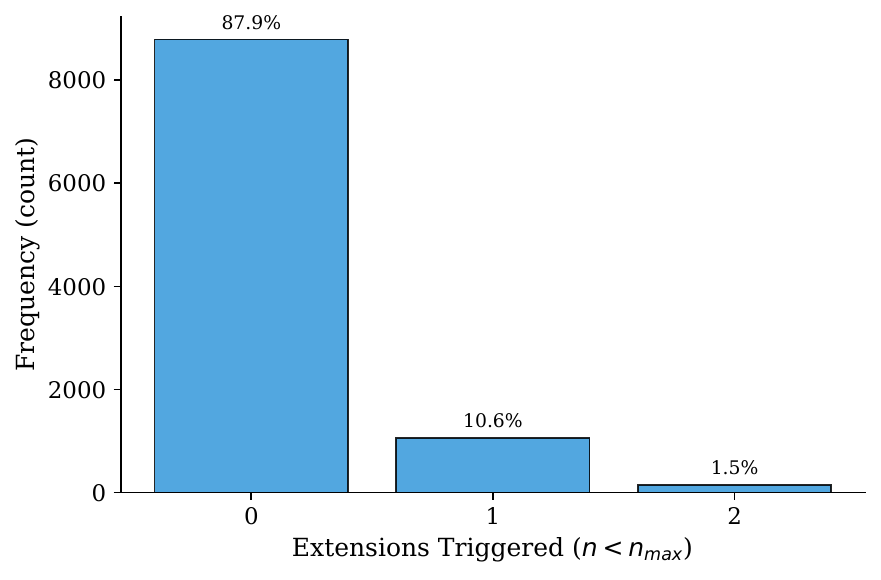}
  \caption{Frequency Distribution of Signal Extensions Triggered by NPLB System}
  \label{fig:extensions}
\end{figure}

The distribution of signal extensions (Figure~\ref{fig:extensions}) shows that 87.9\% of crossing cycles require no intervention, because extensions are triggered only when VRUs are detected near signal expiration. NPLB did not impact the majority of crossing cycles, mostly granting 1 extension in 10.6\% of crossing cycles and rarely 2 extensions in 1.5\%.

\begin{figure}[h!]
  \centering
  \includegraphics[width=0.7\linewidth]{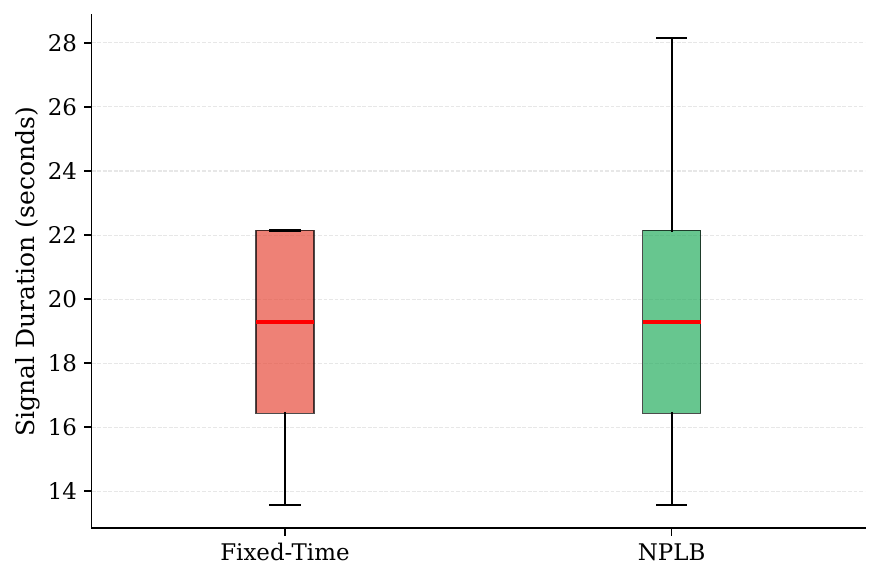}
  \caption{Box Plot Comparing Pedestrian Signal Durations Between Fixed-Time and NPLB Systems}
  \label{fig:signal_times}
\end{figure}

The signal duration analysis (Figure~\ref{fig:signal_times}) demonstrates that NPLB maintains the average signal time achieved by fixed-time signals, the upper bound for fixed-time signals was 22 seconds, while NPLB was 28 seconds. The extension remains modest and only affects 12.1\% of scenarios where VRUs are at risk.

\subsection{Object Detection}
Table~\ref{tab:model_comparison} presents the performance of all five object detection models on the BGVP dataset. Among the YOLO variants, YOLOv12 achieved the highest mAP@0.5 of 0.756, followed closely by YOLOv5 (0.746) and YOLOv11 (0.741). mAP@[0.5:0.95] is a stricter metric which averages performance across multiple IoU thresholds, when evaluating using this stricter metric, YOLOv11 slightly outperforms the other YOLO models with a score of 0.509, compared to 0.506 for YOLOv5 and 0.502 for YOLOv12.

The two-stage Faster R-CNN model achieved a mAP@0.5 of 0.659, performing below all YOLO variants despite its reputation for high accuracy. This suggests that the COCO pre-trained weights may require more extensive fine-tuning on the VRU-specific dataset. The lightweight SSDLite model obtained the lowest performance with a mAP@0.5 of 0.558, which is expected given its design prioritizes computational efficiency over accuracy.

\begin{figure*}[h!]
  \centering
  \includegraphics[width=1\linewidth]{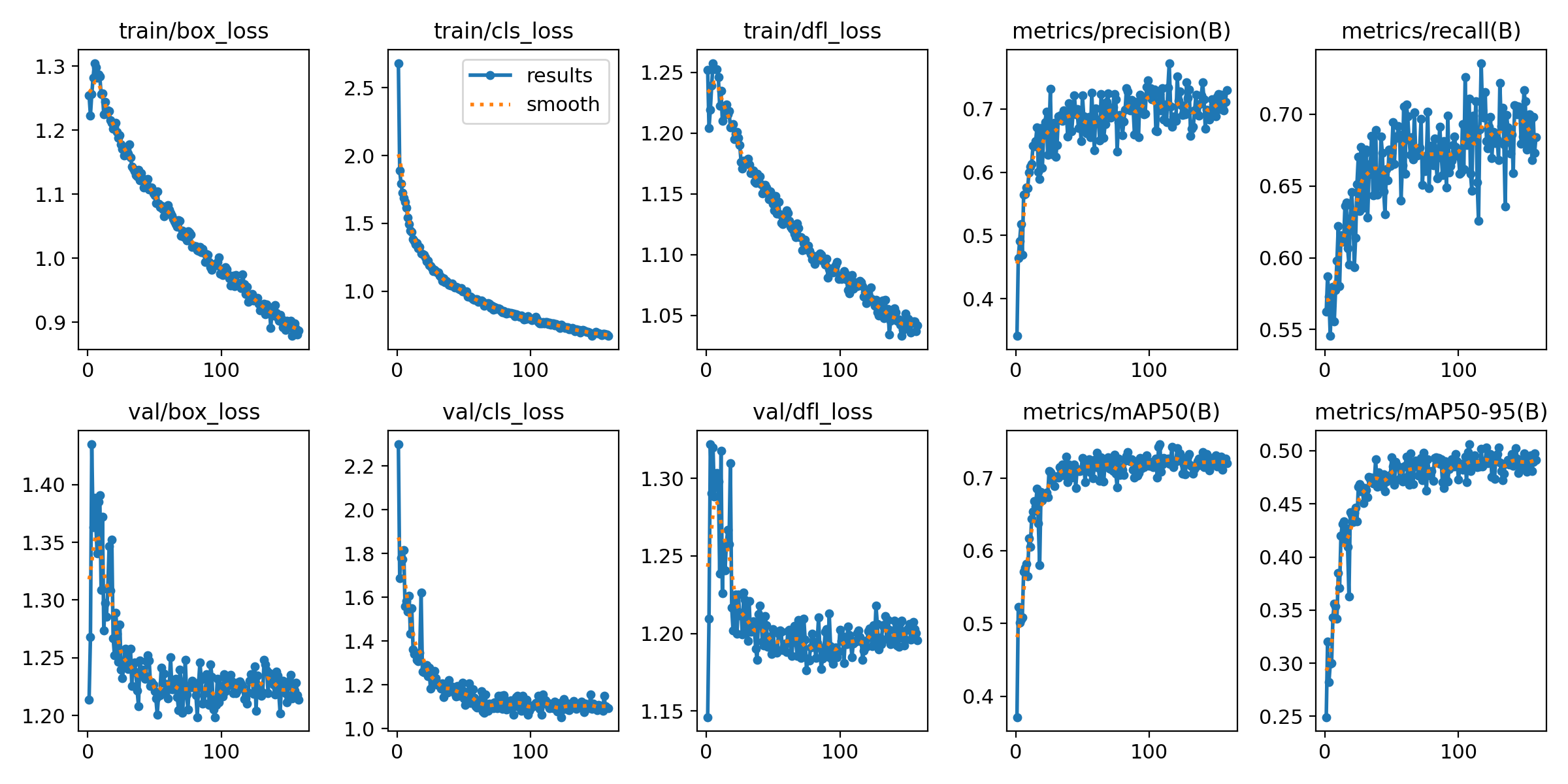}
  \caption{Training and Validation Metrics for YOLOv5 Across 500 Epochs, Showing Convergence of Box Loss, Objectness Loss, Classification Loss, Precision, Recall, and mAP}
  \label{fig:yolov5_results}
\end{figure*}

\begin{figure*}[h!]
  \centering
  \includegraphics[width=1\linewidth]{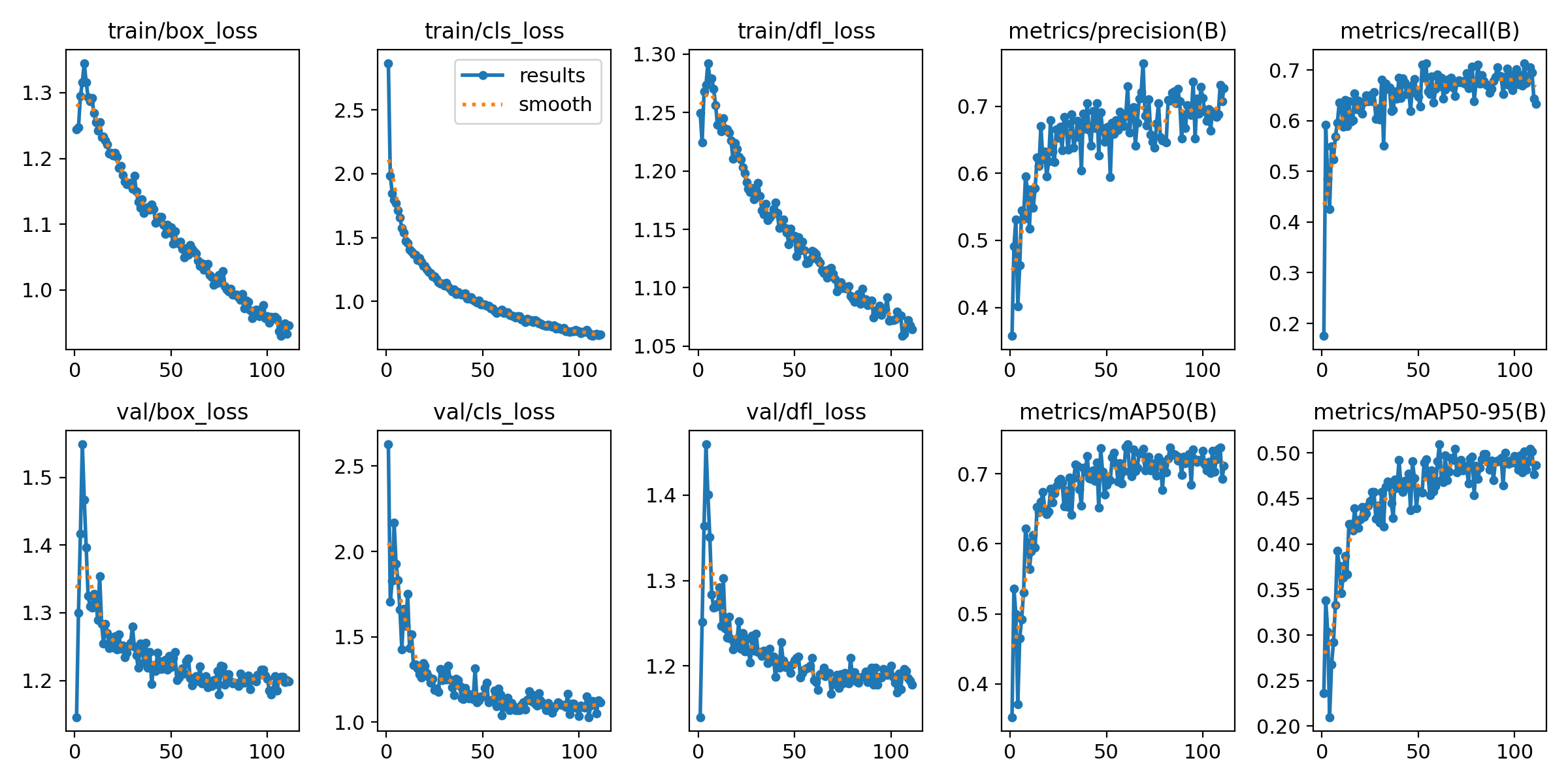}
  \caption{Training and Validation Metrics for YOLOv11 Across 500 Epochs, Showing Convergence of Box Loss, Objectness Loss, Classification Loss, Precision, Recall, and mAP}
  \label{fig:yolov11_results}
\end{figure*}

\begin{figure*}[h!]
  \centering
  \includegraphics[width=1\linewidth]{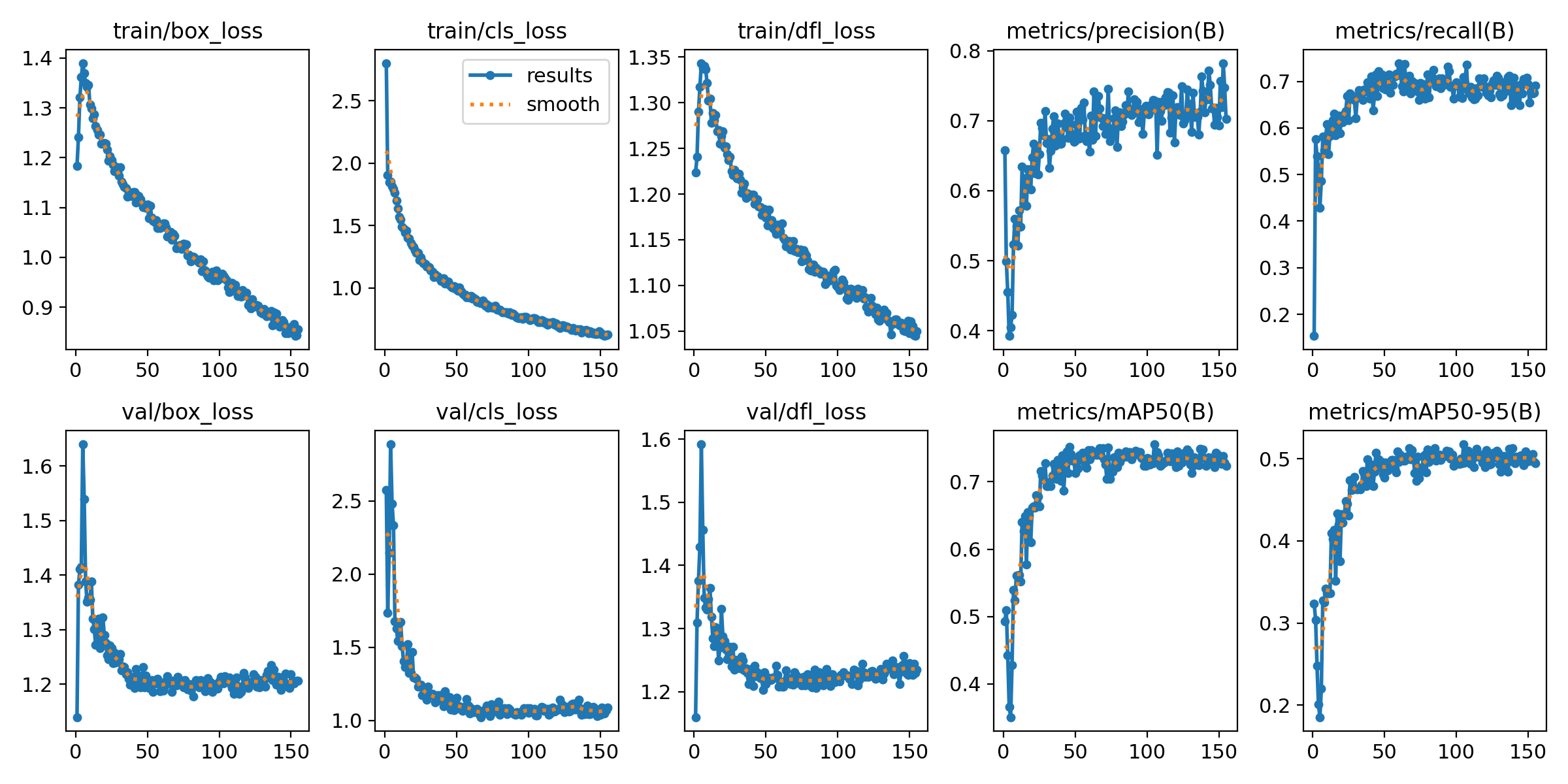}
  \caption{Training and Validation Metrics for YOLOv12 Across 500 Epochs, Showing Convergence of Box Loss, Objectness Loss, Classification Loss, Precision, Recall, and mAP}
  \label{fig:yolov12_results}
\end{figure*}

Overall, the YOLO family demonstrates superior performance for VRU detection on the BGVP dataset, with all three variants achieving mAP@0.5 scores above 0.74 as shown in Figures~\ref{fig:yolov5_results},~\ref{fig:yolov11_results}, and~\ref{fig:yolov12_results}. The minimal performance differences between YOLOv5, YOLOv11, YOLOv12 suggest that any of these architectures would be suitable for deployment in the traffic signal control system.

\begin{table}[h!]
\centering
\caption{Comparison of Object Detection Models}
\begin{tabular}{lcccc}
\hline
\textbf{Model} & \textbf{mAP@0.5} & \textbf{mAP@[0.5:0.95]} \\
\hline
YOLOv12          & 0.756      & 0.502 \\
YOLOv5           & 0.746      & 0.506 \\
YOLOv11          & 0.741      & 0.509 \\
Faster R-CNN     & 0.659      & 0.355  \\
SSDLite (MobileNetV3) & 0.558 & 0.329 \\
\hline
\end{tabular}
\label{tab:model_comparison}
\end{table}

\section{Discussion and Conclusion}
YOLO architectures perform better than other models described in Table~\ref{tab:model_comparison} for the VRU detection task despite smaller size and faster inference. Therefore, Single-stage detectors like YOLO rank as competitive contestants against two-stage architectures in pedestrian safety applications. The 71.4\% stranding reduction achieved by NPLB comes at little disruption to traffic flow, as seen in Figure~\ref{fig:extensions}. The key to this solution is selectivity: by triggering extensions only when VRUs approach signal expiration, the system intervenes rarely but effectively. 

Current transportation infrastructure may unintentionally favor able-bodied pedestrians. NPLB addresses this equity gap by accommodating Vulnerable Road Users. However, camera-based monitoring may raise privacy concerns for stakeholders — posing community engagement and transparent data policies as critical.

Deploying NPLB at an existing crosswalk would require the following: On the hardware side: (1) a weatherproof camera with sufficient resolution positioned to cover the full crosswalk width, (2) An edge compute device capable of running YOLOv12 inference in real time, (3) A hardware interface to the existing traffic signal controller (typically via a relay or controller area network connection). On the software side: (1) The NPLB system with the fine-tuned YOLOv12 weights and ByteTrack integration, (2) A signal controller interface layer compatible with the intersection's existing management system.

\subsection{Future Work and Limitations}
Our Monte Carlo evaluation worked under idealized assumptions: (1) consistent lighting, (2) legal crossing behavior, and (3) all detected pedestrians are crossing. Real-world deployment will encounter visual occlusions caused by vehicles, weather conditions and edge cases that our simulations do not capture.

The fixed threshold parameters ($\tau_e$,  $\tau_t$) may need to be adjusted based on intersection length, the demographics of pedestrians and the traffic patterns. A more robust production system could potentially learn thresholds adaptively from local data rather than relying on pre-defined constants.
\balance
Future work should explore developing a dedicated annotated video dataset of real-world crosswalk interactions with VRU demographic labels, which would enable end-to-end video-based evaluation of NPLB and reduce reliance on simulation.

\bibliographystyle{ACM-Reference-Format}
\bibliography{references}

\end{document}